\documentclass[a4paper]{llncs}

\usepackage{array}

\usepackage{amsmath,amsfonts,amssymb}
\usepackage{graphicx}
\usepackage{subfigure}

\newcommand{\coeffs}{\beta}
\newcommand{\sourcecodeurl}{\url{https://github.com/blaschko/ksupport}}

\begin{document}
\title{A Note on $k$-support Norm Regularized Risk Minimization}
\author{Matthew B.\ Blaschko}
\institute{\'{E}cole Centrale Paris \\ Grande Voie des Vignes\\92295 Ch\^{a}tenay-Malabry, France\\ \email{matthew.blaschko@inria.fr} }

\maketitle

\begin{abstract}
The $k$-support norm has been recently introduced to perform correlated sparsity regularization~\cite{DBLP:conf/nips/ArgyriouFS12}.  Although Argyriou et al.\ only reported experiments using squared loss, here we apply it to several other commonly used settings resulting in novel machine learning algorithms with interesting and familiar limit cases.  Source code for the algorithms described here is available from \sourcecodeurl.
\end{abstract}

\section{The $k$-support Norm}

The $k$-support norm is the gauge function associated with the convex set
\begin{equation}\label{eq:ksupportConvRelaxationFormulation}
\operatorname{conv}\{ \coeffs \ | \ \|\coeffs\|_0 \leq k, \|\coeffs\|_2 \leq 1 \}.
\end{equation}
It can be computed as
\begin{equation}\label{eq:k-supportNorm}
\|\coeffs\|_{k}^{sp} = \left( \sum_{i=1}^{k-r-1} (
  |\coeffs|_{i}^{\downarrow})^2 + \frac{1}{r+1} \left(\sum_{i=k-r}^{d}
    |\coeffs|_{i}^{\downarrow} \right)^2 \right)^{\frac{1}{2}}
\end{equation}
where $|\coeffs|_{i}^{\downarrow}$ is the $i$th largest element of the
vector and $r$ is the unique integer in $\{0,\dots,k-1\}$ satisfying
\begin{equation}
|\coeffs|_{k-r-1}^{\downarrow} > \frac{1}{r+1} \sum_{i=k-r}^{d}
|\coeffs|_{i}^{\downarrow} \geq |\coeffs|_{k-r}^{\downarrow}.
\end{equation}

We use the following notation here: $X \in \mathbb{R}^{n \times d}$ is a design matrix of $n$ samples each with $d$ dimensions; $y \in \mathbb{R}^n$ is the vector of targets.

In the case that $k = 1$ the $k$-support norm is exactly equivalent to the $\ell_1$ norm.  In the case that $k = d$, where $\coeffs \in \mathbb{R}^d$, the $k$-support norm is equivalent to the $\ell_2$ norm.

We note that for an objective
\begin{equation}
\min_{\coeffs} \lambda \| \coeffs \|_{k}^{sp} + f(\coeffs,X,y)
\end{equation}
with some loss function $f(\cdot,\cdot,\cdot)$, when $k=d$, this is equivalent to
\begin{equation}
\min_{\coeffs} \lambda \| \coeffs \|_2 + f(\coeffs,X,y)
\end{equation}
rather than the familiar squared $\ell_2$ regularizer.  However, for any $\lambda$ there exists some $\tilde{\lambda}$ such that
\begin{equation}
\arg\min_{\coeffs} \lambda \| \coeffs \|_2 + f(\coeffs,X,y) = \arg\min_{\coeffs} \tilde{\lambda} \| \coeffs \|_2^{2} + f(\coeffs,X,y) .
\end{equation}
This can be easily seen by noting that the objectives are the Lagrangians of constrained minimization problems that minimize $f$ subject to the equivalent constraints $\| \coeffs \|_2 \leq B$ and $\| \coeffs \|_2^2 \leq B^2$, respectively, for some $B \in \mathbb{R}_{+}$.

\section{Squared Loss}

If we use Nesterov's accelerated method (a first-order proximal algorithm) for optimization as suggested in~\cite{DBLP:conf/nips/ArgyriouFS12}, a given implementation of $k$-support regularized risk requires a function that computes the loss $f$, a function that computes the gradient of the loss function $\frac{\partial f}{\partial \coeffs}$, and the Lipschitz constant $L$ for $\frac{\partial f}{\partial \coeffs}$.  We assume that $f$ is convex and differentiable everywhere and that $L$ is finite.

For the squared loss:
\begin{eqnarray}
f_2(\coeffs, X, y) &=& \| X \coeffs - y \|^{2} \\
\frac{\partial f_2}{\partial \coeffs} &=& 2 X^{T} X \coeffs - 2 X^{T} y \\
L_2 &=& 2 \gamma
\end{eqnarray}
where $\gamma$ is the largest eigenvalue of $X^{T} X$.

The objective function
\begin{equation}
\lambda \|\coeffs\|_k^{sp} + \| X \coeffs - y \|^{2}
\end{equation}
clearly has the lasso~\cite{tibshirani96regression} and ridge regression~\cite{Tikhonov1963} as special cases when $k=1$ and $k=d$, respectively. Argyriou et al.~\cite{DBLP:conf/nips/ArgyriouFS12} have previously discussed the relationship to the elastic net~\cite{Zou05regularization}.
The $k$-support norm with squared loss has been shown to give good results on fMRI data~\cite{GkirtzouISBI2013}.

\section{One Sided Squared Loss}

While we have previously assumed that $y \in \mathbb{R}^n$, here we will assume we are dealing with the binary classification case where $y \in \{ -1, +1 \}^n$.  One sided squared loss simply computes the squared loss when a margin is violated, and zero otherwise.
\begin{eqnarray}
f_{2-}(\coeffs, X, y)  &=& \sum_{i=1}^n \left( \max \{1 - y_i \langle \coeffs, x_i \rangle,0\} \right)^{2}\\
\frac{\partial f_{2-}}{\partial \coeffs} &=& \sum_{i=1}^{n} \begin{cases} 0 & \text{if } y_i \langle \coeffs, x_i \rangle > 1 \\ 2 \langle \coeffs, x_i \rangle x_i - 2 y_i x_i & \text{if } y_i \langle \coeffs, x_i \rangle \leq 1 \end{cases}\\
L_{2-} &=& 2 \gamma.
\end{eqnarray}
One sided squared loss has been considered, for example, in~\cite{Chapelle:2007:TSV:1246422.1246423}.

\section{Hinge Loss}\label{sec:HingeLoss}

Hinge loss is not differentiable, so we apply a Huber approximation to hinge loss~\cite{Chapelle:2007:TSV:1246422.1246423}.\footnote{Although it is perhaps more natural to incorporate non-differentiable losses with the k-support regularizer in a proximal splitting approach, we have arbitrarily closely approximated non-differentiable losses by differentiable ones for the sake of uniformity of presentation and software implementation.}  The Huber parameter is denoted $h$:
\begin{eqnarray}
f_h(\coeffs, X, y) &=& \sum_{i=1}^{n} \begin{cases} 0 & \text{if } y_i \langle \coeffs, x_i \rangle > 1+h \\ \frac{(1 + h - y_i \langle \coeffs, x_i \rangle )^2}{4h} & \text{if } |1-y_i \langle \coeffs, x_i \rangle | \leq h \\ 1 - y_i \langle \coeffs, x_i \rangle & \text{if } y_i \langle \coeffs , x_i \rangle < 1-h \end{cases} \\
\frac{\partial f_h}{\partial \coeffs} &=& \sum_{i=1}^{n} \begin{cases} 0 & \text{if } y_i \langle \coeffs , x_i \rangle >1+h \\ \frac{\langle \coeffs , x_i \rangle x_i - (1+h) y_i x_i}{2h} & \text{if } |1-y_i \langle \coeffs, x_i \rangle | \leq h \\ - y_i x_i & \text{if } y_i \langle \coeffs , x_i \rangle < 1-h \end{cases}  \\
L_2 &=& \frac{\gamma}{2 h}
\end{eqnarray}
where $\gamma$ is as before the largest eigenvalue of $X^{T} X$.  We note that the Lipschitz constant is in a sense conservative in that it grows with the inverse of $h$, while we might expect a smaller fraction of the data to actually fall within the quadratic portion of the data.  Nevertheless for $h$ not too small, we have not observed any convergence issues with Nesterov's accelerated method.  While a small value of $h$ may be desirable in a kernelized setting, here we desire Hinge loss not for sparsity of a dual coefficient vector (indeed the $k$-support norm does not admit a representer theorem~\cite{ArgyriouetalJMLR2009}), but rather that the loss not grow more than linearly while remaining convex.  In other words, we use the hinge loss primarily for its increased robustness over other losses such as (one-sided) squared loss.

The limit cases are the support vector machine (SVM)~\cite{Cortes:1995:SN:218919.218929} when $k = d$ and the $\ell_1$ regularized SVM~\cite{NIPS2003_AA07} when $k=1$.  The $k$-support regularized SVM can be seen as an alternative to the elastic net regularized SVM~\cite{WangStatisticaSinica2006}, but with a tighter convex relaxation to correlated sparsity (Equation~\eqref{eq:ksupportConvRelaxationFormulation}).

\section{Logistic Loss}

Logistic loss is derived from logistic regression, and its minimization is equivalent to logistic regression in the case that it is unregularized~\cite{HastieESL2009}.
\begin{eqnarray}
f_{\log}(\coeffs, X, y) &=& \sum_{i=1}^{n} \log\left( 1 + e^{-y_i \langle \coeffs, x_i \rangle} \right)\\
\frac{\partial f_{\log}}{\partial \coeffs} &=& - \sum_{i=1}^{n} \frac{e^{-y_i \langle \coeffs, x_i \rangle}}{1 + e^{-y_i \langle \coeffs, x_i\rangle}} y_i x_i \\
L_{\log} &=& \frac{\gamma}{4}
\end{eqnarray}
where the Lipschitz constant has a factor $\frac{1}{4}$ from the Lipschitz constant of the sigmoid in $\frac{\partial f_{\log}}{\partial \coeffs}$.  $k$-support regularized regression specializes to previously used regularized logistic regression objectives~\cite{Ng:2004:FSL:1015330.1015435} when $k=1$ or $k=d$.

\section{Exponential Loss}

Exponential loss is known primarily through its use in AdaBoost.M1~\cite{FreundS96,HastieESL2009}. 
\begin{eqnarray}
f_{\exp}(\coeffs, X, y) &=& \sum_{i=1}^{n} e^{-y_i \langle \coeffs, x_i \rangle}\\
\frac{\partial f_{\exp}}{\partial \coeffs} &=& - \sum_{i=1}^{n} e^{-y_i \langle \coeffs, x_i \rangle} y_i x_i
\end{eqnarray}
Here, the loss is not globally Lipschitz continuous.  However, one may attempt to estimate a sufficiently large constant if one were to apply learning with the $k$-support norm and Nesterov's accelerated method (we have simply used a relatively conservative $50 \times \gamma$ in the experiments reported in Section~\ref{sec:ExperimentalResults}).  As exponential loss is highly degenerate in the presence of label noise (essentially for the same reason that it is not globally Lipschitz continuous), this is likely of limited utility in real-world applications.  We have included this loss here primarily for completeness, and have not explored any other optimization strategies.

\section{$\varepsilon$-insensitive Loss and Huber Smoothed Absolute Loss}

\begin{figure}
    \centering
    \subfigure[$\varepsilon=2$ gives an insensitive region around the correct regression value.]{
        \includegraphics[width=0.6\columnwidth]{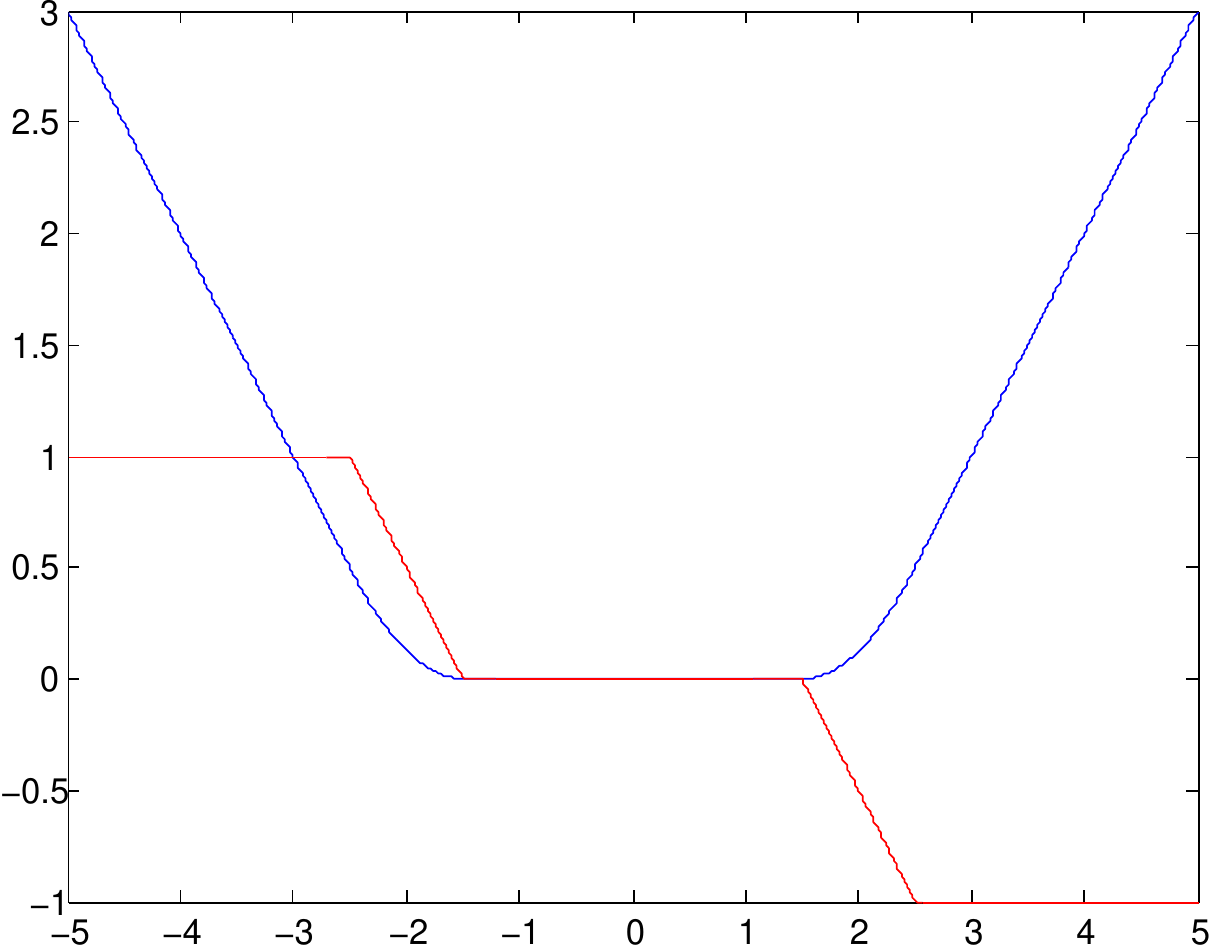}}
\subfigure[The special case that $\varepsilon = 0$ results in a Huber smoothed absolute loss.]{
        \includegraphics[width=0.6\columnwidth]{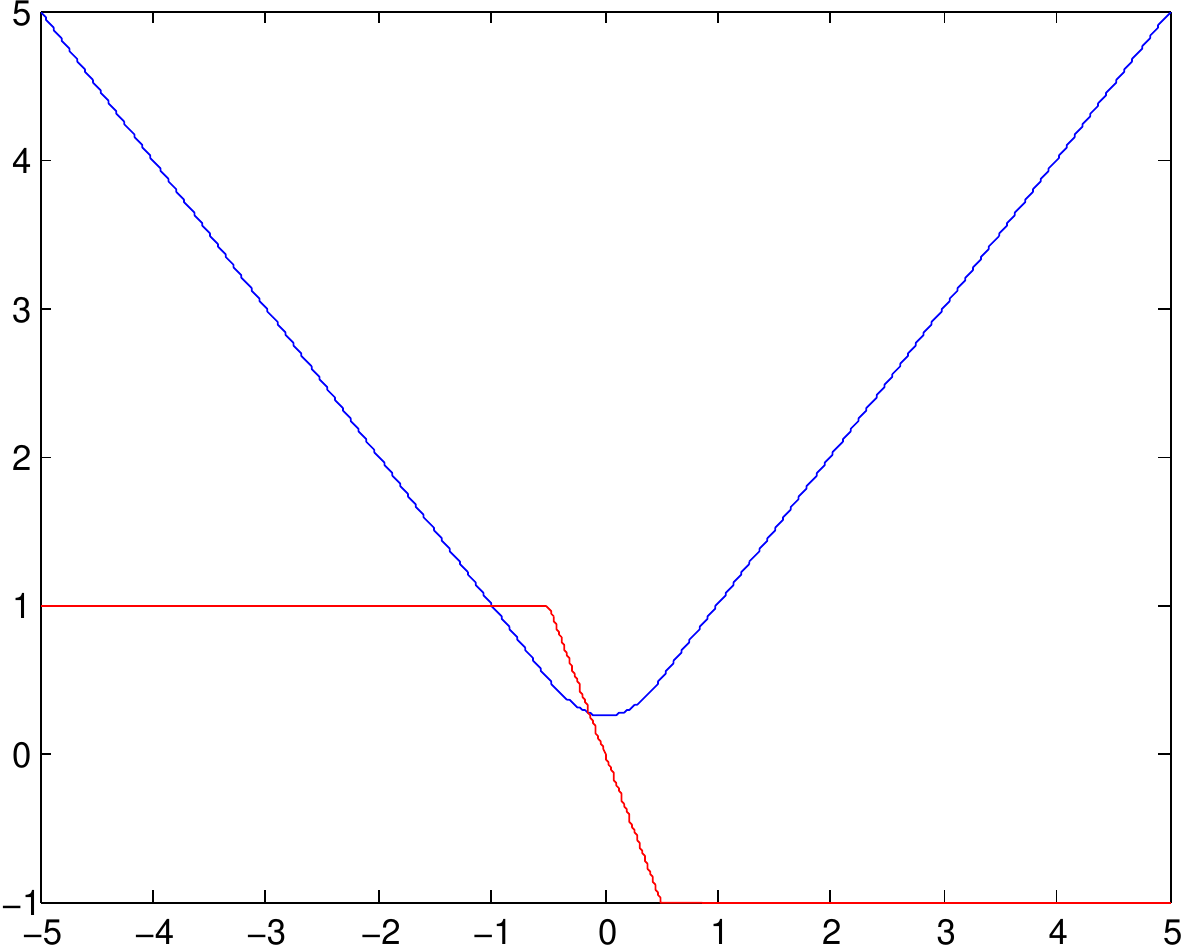}
}
      \caption{Huber smoothed $\varepsilon$-insensitive loss.  On the horizontal axis is $y_i - \langle \coeffs, x_i \rangle$ while the vertical axis plots $f_{\varepsilon}$ in blue, and $\frac{\partial f_{\varepsilon}}{\partial \coeffs}$ in red.  In both plots $h = \frac{1}{2}$.}\label{fig:HuberEpsLoss}
\end{figure}

$\varepsilon$-insensitive loss is defined to be~\cite{Vapnik:1995:NSL:211359}: 
\begin{equation}
| y_i - \langle \coeffs, x_i \rangle |_{\varepsilon} := \max \{ 0, |y - \langle \coeffs, x_i \rangle | - \varepsilon \}
\end{equation}
for some parameter $\varepsilon \geq 0$.  While $\varepsilon$ has an important role in the sparsity of the dual representation for support vector regression~\cite{Vapnik:1995:NSL:211359}, that role is not required in the primal.  As with hinge loss, we use Huber smoothing to guarantee differentiability.
\begin{eqnarray} \label{eq:epsInsLoss}
f_{\varepsilon}(\coeffs,X,y) &=& \sum_{i=1}^{n} \begin{cases} 0 & \text{if } y_i - \langle \coeffs, x_i \rangle > -\varepsilon + h \\ \frac{(y_i - \langle \coeffs, x_i \rangle + \varepsilon - h)^2}{4h} & \text{if } | y_i - \langle \coeffs, x_i \rangle + \varepsilon | \leq h \\ -y_i + \langle \coeffs, x_i \rangle - \varepsilon & \text{if } y_i - \langle \coeffs, x_i \rangle < -\varepsilon - h \end{cases} \\
&+& \sum_{i=1}^{n} \begin{cases} 0 & \text{if } y_i - \langle \coeffs, x_i \rangle < \varepsilon - h \\ \frac{(y_i - \langle \coeffs, x_i \rangle - \varepsilon + h)^2}{4h} & \text{if } | y_i - \langle \coeffs, x_i \rangle - \varepsilon | \leq h \\ y_i - \langle \coeffs, x_i \rangle - \varepsilon & \text{if } y_i - \langle \coeffs, x_i \rangle > \varepsilon + h \end{cases}    \nonumber \\
\frac{\partial f_{\varepsilon}}{\partial \coeffs} &=& \sum_{i=1}^{n} \begin{cases} 0 & \text{if } y_i - \langle \coeffs, x_i \rangle > -\varepsilon + h \\ \frac{ \langle \coeffs, x_i \rangle x_i + (-\varepsilon + h - y_i) x_i}{2h} & \text{if } | y_i - \langle \coeffs, x_i \rangle + \varepsilon | \leq h \\ x_i  & \text{if } y_i - \langle \coeffs, x_i \rangle < -\varepsilon - h \end{cases} \\
&+& \sum_{i=1}^{n} \begin{cases} 0 & \text{if } y_i - \langle \coeffs, x_i \rangle < \varepsilon - h \\ \frac{\langle \coeffs, x_i \rangle x_i + (\varepsilon - h - y_i) x_i}{2h} & \text{if } | y_i - \langle \coeffs, x_i \rangle - \varepsilon | \leq h \\ - x_i & \text{if } y_i - \langle \coeffs, x_i \rangle > \varepsilon + h \end{cases}    \nonumber \\
L_{\varepsilon} &=& \frac{\gamma}{h}
\end{eqnarray}
Here we have decomposed the $\varepsilon$ insensitive loss into two hinge components to emphasize the relationship to Huber smoothed hinge loss (cf.\ Section~\ref{sec:HingeLoss}).  A plot of the loss and its gradient is shown in Figure~\ref{fig:HuberEpsLoss}.
In the case that $\varepsilon=0$ we get a Huber smoothed absolute loss function as a special case (denoted $f_{\operatorname{abs}}$ in the sequel), and the curvature of the loss function at $y_i - \langle \coeffs,x_i \rangle = 0$ is doubled, therefore the Lipschitz constant is double that of the one sided hinge loss.

In the case that $k=d$, we recover the special case of $\varepsilon$-support vector regression ($\varepsilon$-SVR)~\cite{Scholkopf:2001:LKS:559923}. 
If we set $k=1$ we get an $\ell_1$ regularized variant of $\varepsilon$-SVR.  In the case that $\varepsilon=0$ this $\ell_1$ regularized variant is equivalent to regularized least absolute deviations regression~\cite{Wang:2006:RLA:1193207.1193371}. In Equation~\eqref{eq:epsInsLoss}, $\varepsilon<0$ is equivalent to $\varepsilon>0$ but with a constant value added to the loss everywhere, i.e.\ the minimizer is the same.

\section{Experiments}\label{sec:ExperimentalResults}

\begin{table}[t]
\caption{Accuracies for each method and regularizer.  See text for the experimental setting.  The $k$-support norm achieved higher acuracies on average for all loss functions.}\label{tab:ToyAccuracies}
\resizebox{\textwidth}{!}{
{\setlength{\tabcolsep}{0.6em}%
\begin{tabular}{l|ccccccc}
 & $f_2$ & $f_{2-}$ & $f_h$ & $f_{\log}$ & $f_{\exp}$ & $f_{\operatorname{abs}}$ & $f_{\varepsilon}$ \\
\hline
$\| \coeffs \|_k^{sp}$ & $0.883 \pm 0.058$  &  $0.883 \pm 0.058$  &  $0.890 \pm 0.057$  &  $0.889 \pm 0.056$  &  $0.888 \pm 0.060$  &  $0.889 \pm 0.065$  &  $0.886 \pm 0.062$\\
$\| \coeffs \|_1$      & $0.870 \pm 0.062$  &  $0.870 \pm 0.062$  &  $0.868 \pm 0.069$  &  $0.872 \pm 0.063$  &  $0.876 \pm 0.065$  &  $0.870 \pm 0.077$  &  $0.879 \pm 0.059$\\
$\| \coeffs \|_2$      & $0.871 \pm 0.071$  &  $0.871 \pm 0.071$  &  $0.872 \pm 0.065$  &  $0.872 \pm 0.066$  &  $0.870 \pm 0.067$  &  $0.867 \pm 0.071$  &  $0.872 \pm 0.063$\\
\end{tabular}
}}
\end{table}


\begin{table}[t]
\caption{Mean squared errors (MSE) for each method and regularizer.  See text for the experimental setting.  $f_2$, $f_{2-}$, $f_{\operatorname{abs}}$, and $f_{\varepsilon}$ achieved the lowest MSEs with the $k$-support norm regularizer giving best results on average.}\label{tab:ToyMSE}
\resizebox{\textwidth}{!}{
{\setlength{\tabcolsep}{0.6em}%
\begin{tabular}{l|ccccccc}
 & $f_2$ & $f_{2-}$ & $f_h$ & $f_{\log}$ & $f_{\exp}$ & $f_{\operatorname{abs}}$ & $f_{\varepsilon}$ \\
\hline
$\| \coeffs \|_k^{sp}$ & $1.21e2 \pm 4.89e1$  &  $1.21e2 \pm 4.89e1$  &  $1.78e2 \pm 1.00e2$  &  $3.33e3 \pm 5.39e3$  &  $1.59e3 \pm 2.89e3$  &  $1.25e2 \pm 5.41e1$  & $2.21e2 \pm 1.51e1$\\
$\| \coeffs \|_1$      & $1.25e2 \pm 4.81e1$  &  $1.25e2 \pm 4.81e1$  &  $2.21e2 \pm 9.63e1$  &  $1.13e4 \pm 9.89e3$  &  $6.16e3 \pm 4.82e3$  &  $1.48e2 \pm 1.76e2$  & $2.16e2 \pm 1.66e1$\\
$\| \coeffs \|_2$      & $1.49e2 \pm 4.75e1$  &  $1.49e2 \pm 4.74e1$  &  $1.81e2 \pm 7.66e1$  &  $4.18e3 \pm 8.00e3$  &  $3.08e3 \pm 4.88e3$  &  $1.50e2 \pm 5.34e1$  & $2.25e2 \pm 1.56e1$\\
\end{tabular}
}}
\end{table}

We have applied each of the algorithms above to a toy classification problem conceptually similar to that reported in~\cite{DBLP:conf/nips/ArgyriouFS12}.
In all cases, we perform model selection for $k \in \{1, \dots , d\}$ and $\lambda = 10^i, \ i \in \{-15,\dots , 5\}$.  We compare additionally to the special fixed cases $k=1$ and $k=d$ corresponding to $\ell_1$ and $\ell_2$ regularization, respectively.

Output labels were generated randomly with equal probability.  The first 15 dimensions were set by multiplying the label by a fixed vector of 15 samples from a zero mean Gaussian and adding Gaussian noise (i.e.\ a noisy signal is contained in the first 15 dimensions).  The subsequent 50 dimensions were set to zero mean Gaussian noise (i.e.\ the subsequent dimensions contain no signal and should be ignored).  50 samples were used for training, 50 for validation, and 250 for testing.  Table~\ref{tab:ToyAccuracies} gives the mean accuracies for each method across 20 random problem instances, while Table~\ref{tab:ToyMSE} gives the mean squared error (MSE).  For $\varepsilon$-insensitive loss we arbitrarily set $\varepsilon=1$.  For all methods with a Huber smothing parameter, we set $h=\frac{1}{10}$.

It should be noted that several of the methods employed here for classification were developed for regression (squared loss, absolute loss, and $\varepsilon$-insensitive loss).  The experiments performed here were done primarily to validate their correct implementation.

\section{Conclusions}

We have described and implemented a large number of loss functions for non-differentiably regularized risk optimization with proximal splitting methods.  These loss functions in combination with the $k$-support norm yield a large number of learning algorithms proposed in the literature as special cases.  Assuming zero model error, each of these loss functions is sufficient to yield a statistically consistent algorithm\footnote{Huber smoothed $\varepsilon$-insensitive loss requires that $\varepsilon - h < 1$ for consistency in the binary classification setting, $y_i \in \{-1, +1 \}$.} (provided regularization goes to zero at a sufficient rate as the number of samples goes to infinity)~\cite[Theorem 4]{bjm-ccrb-05}.  However, their finite sample behavior varies substantially.  We hope that their implementation and description in a common framework will facilitate their analysis and employment in machine learning studies and applications.

\bibliographystyle{splncs03}
\bibliography{bibliography}

\end{document}